\documentclass[11pt,letterpaper]{article}
\usepackage[letterpaper]{geometry}
\usepackage{acl2012}
\usepackage{times}
\usepackage{latexsym}
\usepackage{amsmath}
\usepackage{multirow}
\usepackage{url}
\usepackage{graphicx}
\makeatletter
\newcommand{\@BIBLABEL}{\@emptybiblabel}
\newcommand{\@emptybiblabel}[1]{}
\makeatother
\usepackage[hidelinks]{hyperref}

\usepackage{pifont}% http://ctan.org/pkg/pifont
\newcommand{\cmark}{\ding{51}}%
\newcommand{\xmark}{\ding{55}}%
\makeatletter
\renewcommand*{\p@section}{\S\,}
\renewcommand*{\p@subsection}{\S\,}

\title{Knowledge as a Teacher: \\Knowledge-Guided Structural Attention Networks}

\author{Yun-Nung Chen$^{\star}$\quad Dilek Hakkani-T\"{u}r$^{\dagger}$\quad Gokhan Tur$^{\dagger}$\\{\bf Asli Celikyilmaz$^{\ddagger}$ \quad Jianfeng Gao$^{\ddagger}$ \quad Li Deng$^{\ddagger}$}\\
  $^{\star}$National Taiwan University, Taipei, Taiwan \\
  $^{\dagger}$Google Research, Mountain View, CA \\
  $^{\ddagger}$Microsoft Research, Redmond, WA \\
  {\tt \{$^{\star}$y.v.chen, $^{\dagger}$dilek, $^{\dagger}$gokhan\}@ieee.org}\\ 
  {\tt \{$^{\ddagger}$aslicel, $^{\ddagger}$jfgao, $^{\ddagger}$deng\}@microsoft.com}\\}

\date{}

\begin{document}
\maketitle
\begin{abstract}
Natural language understanding (NLU) is a core component of a spoken dialogue system. Recently recurrent neural networks (RNN) obtained strong results on NLU due to their superior ability of preserving sequential information over time. Traditionally, the NLU module tags semantic slots for utterances considering their flat structures, as the underlying RNN structure is a linear chain. However, natural language exhibits linguistic properties that provide rich, structured information for better understanding. This paper introduces a novel model, knowledge-guided structural attention networks (K-SAN), a generalization of RNN to additionally incorporate non-flat network topologies guided by prior knowledge. There are two characteristics: 1) important substructures can be captured from small training data, allowing the model to generalize to previously unseen test data; 2) the model automatically figures out the salient substructures that are essential to predict the semantic tags of the given sentences, so that the understanding performance can be improved. The experiments on the benchmark Air Travel Information System (ATIS) data show that the proposed K-SAN architecture can effectively extract salient knowledge from substructures with an attention mechanism, and outperform the performance of the state-of-the-art neural network based frameworks.
%\footnote{The code and pre-trained model will be released.}.
\end{abstract}

\section{Introduction}

In the past decade, goal-oriented spoken dialogue systems (SDS), such as the virtual personal assistants Microsoft's Cortana and Apple's Siri, are being incorporated in various devices and allow users to speak to systems freely in order to finish tasks more efficiently.
A key component of these conversational systems is the natural language understanding (NLU) module—-it refers to the targeted understanding of human speech directed at machines~\cite{tur2011spoken}.
The goal of such ``targeted'' understanding is to convert the recognized user speech into a task-specific semantic representation of the user's intention, at each turn, that aligns with the back-end knowledge and action sources for task completion.
The dialogue manager then interprets the semantics of the user's request and associated back-end results, and decides the most appropriate system action, by exploiting semantic context and user specific meta-information, such as geo-location and personal preferences~\cite{mctear2004spoken,rudnicky1999agenda}.

\begin{figure}[t]
	\centering
    \includegraphics[width=0.9\linewidth]{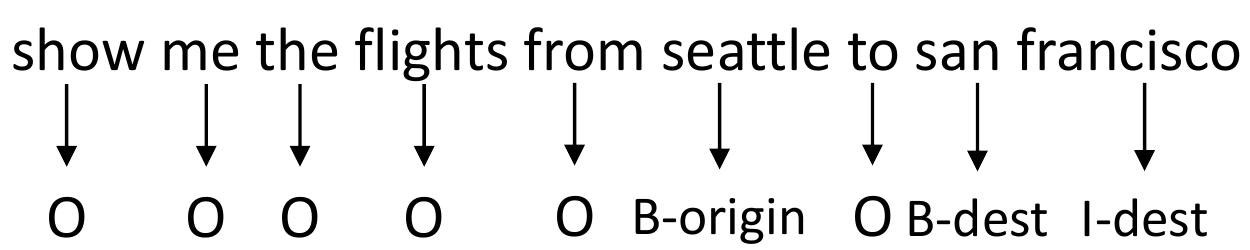}
    \vspace{-2mm}
    \caption{An example utterance annotated with its semantic slots in the IOB format (S).}
\vspace{-4mm}
    \label{fig:iob}
\end{figure}

A typical pipeline of NLU includes:
domain classification, intent determination, and slot filling~\cite{tur2011spoken}.
NLU first decides the domain of user's request given the input utterance, and based on the domain, predicts the intent and fills associated slots corresponding to a domain-specific semantic template.
For example, Figure~\ref{fig:iob} shows a user utterance, ``\textit{show me the flights from seattle to san francisco}'' and its semantic frame, \textsf{find\_flight(origin=``seattle'', dest=``san francisco'')}. %~\cite{chen2015matrix}.
It is easy to see the relationship between the origin city and the destination city in this example, although these do not appear next to each other. 
%\noindent \textbf{Deep Learning for NLU}\\
%The main goal of NLU in task-oriented human-machine conversational systems is to automatically classify the domain of a user utterance along with a domain-specific intent and fill in a set of arguments or ``slots'' forming a semantic frame.
Traditionally, domain detection and intent prediction are framed as utterance classification problems, where several classifiers such as support vector machines and maximum entropy have been employed~\cite{haffner2003optimizing,chelba2003speech,chen2014deriving}.
Then slot filling is framed as a word sequence tagging task, where the IOB (in-out-begin) format is applied for representing slot tags as illustrated in Figure~\ref{fig:iob}, and hidden Markov models (HMM) or conditional random fields (CRF) have been employed for slot tagging~\cite{pieraccini1992speech,wang2005spoken}. %,raymond2007generative}.

%\paragraph{Deep Learning for NLU}
With the advances on deep learning, deep belief networks (DBNs) with deep neural networks (DNNs) have been applied to domain and intent classification tasks~\cite{sarikaya2011deep,tur2012towards,sarikaya2014application}.
Recently, \newcite{ravuri2015recurrent} proposed an RNN architecture for intent determination.
For slot filling, deep learning has been viewed as a feature generator and the neural architecture can be merged with CRFs~\cite{xu2013convolutional}.
\newcite{yao2013recurrent} and \newcite{mesnil2015using} later employed RNNs for sequence labeling in order to perform slot filling.
However, the above studies benefit from large training data without leveraging any existing knowledge.
When tagging sequences RNNs consider them as flat structures, with their underlying linear chain structures, potentially ignoring the structured information typical of natural language sequences.

%\paragraph{Knowledge-Enriched NLU}
Hierarchical structures and semantic relationships contain linguistic characteristics of input word sequences forming sentences, and such information may help interpret their meaning.
Furthermore, prior knowledge would help in the tagging of sequences, especially when dealing with previously unseen sequences~\cite{tur2010left,deoras2013deep}.
%Such knowledge, however, has not been considered during tagging in prior work.
Prior work exploited external web-scale knowledge graphs such as Freebase and Wikipedia for improving NLU~\cite{heck2013leveraging,ma2015knowledge,chen2014deriving}
%,chen2014dynamically},
\newcite{liu2013query} and \newcite{chen2015matrix} proposed approaches that leverage linguistic knowledge encoded in parse trees for language understanding, where the extracted syntactic structural features and semantic dependency features enhance inference model learning, and the model achieves better language understanding performance in various domains.

Even with the emerging paradigm of integrating deep learning and linguistic knowledge for different NLP tasks~\cite{socher2014grounded}, most of the previous work utilized such linguistic knowledge and knowledge bases as additional features as input to neural networks, and then learned the models for tagging sequences.
These feature enrichment based approaches have some possible limitations:
1) poor generalization and 2) error propagation.
Poor generalization comes from the mismatch between knowledge bases and the input data, and then the incorrectly extracted features due to errors in previous processing propagate errors to the neural models.
In order to address the issues and better learn the sequence tagging models, this paper proposes knowledge-guided structural attention networks, K-SAN, a generalization of RNNs that automatically learn the attention guided by external or prior knowledge and generate sentence-based representations specifically for modeling sequence tagging.
The main difference between K-SAN and previous approaches is that knowledge plays the role of a teacher to guide networks where and how much to focus attention considering the whole linguistic structure simultaneously.
Our main contributions are three-fold:
\begin{itemize}
\vspace{-1.5mm}
\item End-to-end learning\\
To our knowledge, this is the first neural network approach that utilizes general knowledge as guidance in an end-to-end fashion, where the model automatically learns important substructures with an attention mechanism.
\vspace{-1.5mm}
\item Generalization for different knowledge\\
There is no required schema of knowledge, and different types of parsing results, such as dependency relations, knowledge graph-specific relations, and parsing output of hand-crafted grammars, can serve as the knowledge guidance in this model.
\vspace{-1.5mm}
\item Efficiency and parallelizability\\
Because the substructures from the input utterance are modeled separately, modeling time may not increase linearly with respect to the number of words in the input sentence.
\vspace{-1.5mm}
\end{itemize}
In the following sections, we empirically show the benefit of K-SAN on the targeted NLU task.

\section{Related Work}
\label{sec:related}

\paragraph{Knowledge-Based Representations}

There is an emerging trend of learning representations at different levels, such as word embeddings~\cite{mikolov2013distributed}, character embeddings~\cite{ling2015finding}, and sentence embeddings~\cite{le2014distributed,huang2013learning}.
%\paragraph{Entity Representation}
In addition to fully unsupervised embedding learning,
knowledge bases have been widely utilized to learn entity embeddings with specific functions or relations~\cite{celikyilmaz2015unsupervised,yang2014embedding}.
Different from prior work, this paper focuses on learning composable substructure embeddings that are informative for understanding.

%\paragraph{Tree-Structured Representation}
Recently linguistic structures are taken into account in the deep learning framework.
\newcite{ma2015dependency} and \newcite{tai2015improved} both proposed dependency-based approaches to combine deep learning and linguistic structures, where the model used tree-based n-grams instead of surface ones to capture knowledge-guided relations for sentence modeling and classification.
\newcite{roth2016neural} utilized lexicalized dependency paths to learn embedding representations for semantic role labeling.
However, the performance of these approaches highly depends on the quality of ``whole'' sentence parsing, and there is no control of degree of attentions on different substructures.
Learning robust representations incorporating whole structures still remains unsolved.
In this paper, we address the limitation by proposing K-SAN to learn robust representations of whole sentences, where the whole representation is composed of the salient substructures in order to avoid error propagation.

\paragraph{Neural Attention and Memory Model}

One of the earliest work with a memory component applied to language processing is memory networks~\cite{weston2015memory,sukhbaatar2015end}, which encode facts into vectors and store them in the memory for question answering (QA).
Following their success, \newcite{xiong2016dynamic} proposed dynamic memory networks (DMN) to additionally capture position and temporality of transitive reasoning steps for different QA tasks.
The idea is to encode important knowledge and store it into memory for future usage with attention mechanisms.
Attention mechanisms allow neural network models to selectively pay attention to specific parts.
There are also various tasks showing the effectiveness of attention mechanisms. %~\cite{}.

However, most previous work focused on the classification or prediction tasks (predicting a single word given a question), and there are few studies for NLU tasks (slot tagging).
Based on the fact that the linguistic or knowledge-based substructures can be treated as prior knowledge to benefit language understanding, this work borrows the idea from memory models to improve NLU.
Unlike the prior NLU work that utilized representations learned from knowledge bases to enrich features of the current sentence, this paper directly learns a sentence representation incorporating memorized substructures with an automatically decided attention mechanism in an end-to-end manner.

\section{Knowledge-Guided Structural Attention Networks (K-SAN)}
\label{sec:model}

\begin{figure*}[t]
\centering
\vspace{-2mm}
\includegraphics[width=\linewidth]{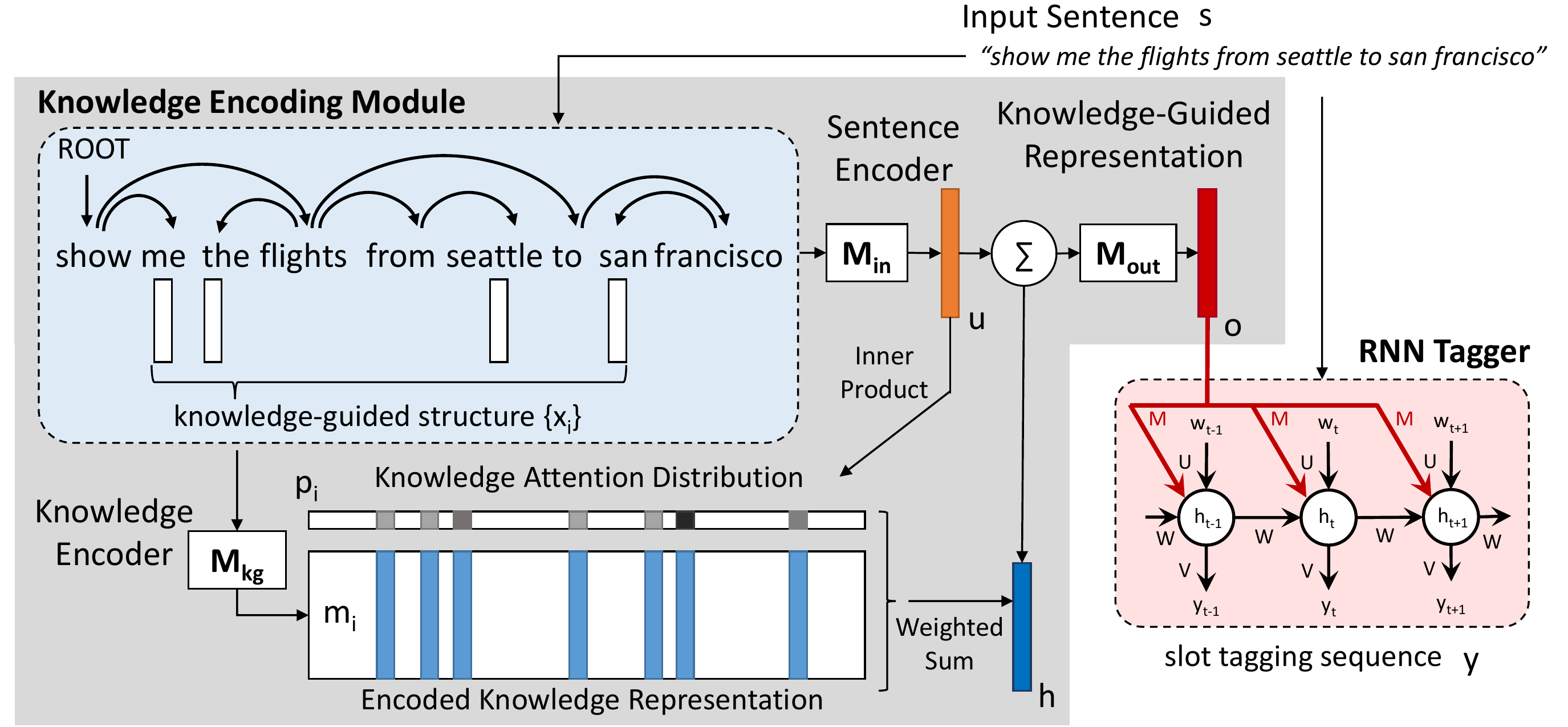}
\vspace{-5mm}
\caption{The illustration of knowledge-guided structural attention networks (K-SAN) for NLU.}
\label{fig:model}
\vspace{-2mm}
\end{figure*}

For the NLU task, given an utterance with a sequence of words/tokens $\vec{s} = w_1, ..., w_T$, our model is to predict corresponding semantic tags $\vec{y} = y_1, ..., y_T$ for each word/token by incorporating knowledge-guided structures.
The proposed model is illustrated in Figure~\ref{fig:model}.
The knowledge encoding module first leverages external knowledge to generate a linguistic structure for the utterance, where a discrete set of knowledge-guided substructures $\{x_i\}$ is encoded into a set of vector representations (\ref{ssec:kg_encoding}).
The model learns the representation for the whole sentence by paying different attention on the substructures (\ref{ssec:arch}).
Then the learned vector encoding the knowledge-guided structure is used for improving the semantic tagger (\ref{sec:tagger}).

\subsection{Knowledge Encoding Module}
\label{ssec:kg_encoding}

\begin{figure}[t]
\centering
\vspace{-2mm}
\includegraphics[width=\linewidth]{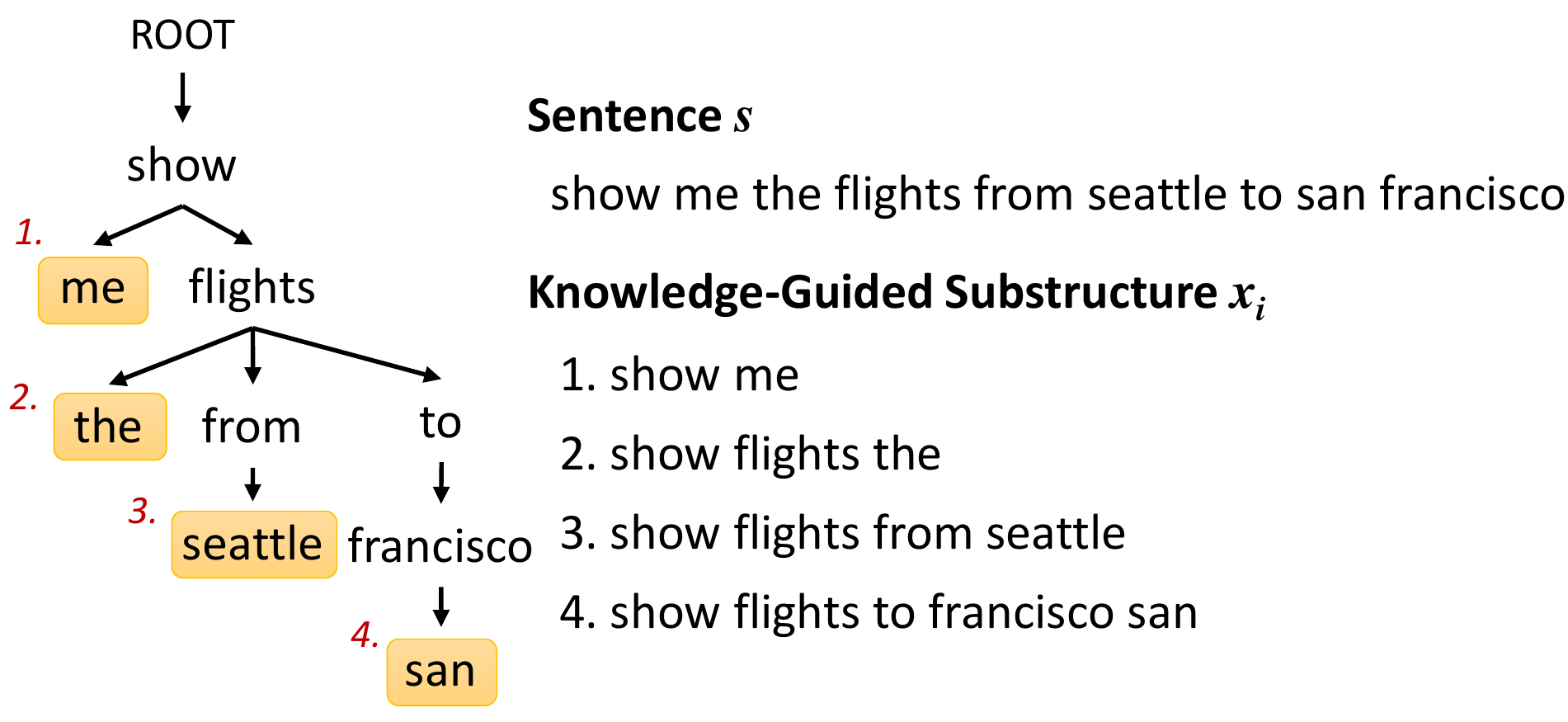}
\vspace{-4mm}
\caption{The knowledge-guided substructures of dependency parsing, $x_i$, on an example sentence $s$.}
\vspace{-2mm}
\label{fig:tree}
\end{figure}

The prior knowledge obtained from external resources, such as dependency relations, knowledge bases, etc., provides richer information to help decide the semantic tags given an input utterance.
This paper takes dependency relations as an example for knowledge encoding, and other structured relations can be applied in the same way.
The input utterance is parsed by a dependency parser, and the substructures are built according to the paths from the root to all leaves~\cite{chen2014fast}.
For example, the dependency parsing of the utterance ``{\it show me the flights from seattle to san francisco}'' is shown in Figure~\ref{fig:tree}, where the associated substructures are obtained from the parsing tree for knowledge encoding.
Here we do not utilize the dependency relation labels in the experiments for better generalization, because the labels may not be always available for different knowledge resources.
Note that the number of substructures may be less than the number of words in the utterance, because non-leaf nodes do not have corresponding substructure in order to reduce the duplicated information in the model.
The top-left component of Figure~\ref{fig:model} illustrates the module for modeling knowledge-guided substructures.

%The model can use different forms of knowledge for guidance, for example, syntactic relations, dependency relations, and relations associated with knowledge graphs.

%\subsubsection{Dependency Tree}
%For the NLU task, our model incorporates a discrete set of history utterances $\{x_i\}$ that are stored in the memory, a current utterance $\vec{s} = w_1, ..., w_T$, and outputs corresponding semantic tags $\vec{y} = y_1, ..., y_T$, where a semantic tag consists intent and slot information.
%The proposed model is illustrated in Figure~\ref{fig:model} and detailed below.

\subsection{Model Architecture}
\label{ssec:arch}
The model embeds all knowledge-guided substructures into a continuous space and stores embeddings of all $x$'s in the knowledge memory.
The representation of the input utterance is then compared with encoded knowledge representations to integrate the carried structure guided by knowledge via an attention mechanism.
Then the knowledge-guided representation of the sentence is taken together with the word sequence for estimating the semantic tags.
Four main procedures are described below.

\paragraph{Encoded Knowledge Representation}
To store the knowledge-guided structure, we convert each substructure (e.g. path starting from the root to the leaf in the dependency tree), $x_i$, into a structure vector $m_i$ with dimension $d$ by embedding the substructure in a continuous space through the knowledge encoding model $M_\text{kg}$.
The input utterance $s$ is also embedded to a vector $u$ with the same dimension through the model $M_\text{in}$.
\begin{eqnarray}
m_i &=& M_\text{kg}(x_i),\\
u &=& M_\text{in}(s).
\end{eqnarray}
We apply the three types for knowledge encoding models, $M_\text{kg}$ and $M_\text{in}$, in order to model multiple words from a substructure $x_i$ or an input sentence $s$ into a vector representation:
1) fully-connected neural networks (NN) with linear activation, 
2) recurrent neural networks (RNN),
and 3) convolutional neural networks (CNN) with a window size 3 and a max-pooling operation.
For example, one of substructures shown in Figure~\ref{fig:tree}, ``{\it show flights seattle from}'', is encoded into a vector embedding.
In the experiments, the weights of $M_\text{kg}$ and $M_\text{in}$ are tied together based on their consistent ability of sequence encoding.
%are tied together to 

\paragraph{Knowledge Attention Distribution}
In the embedding space, we compute the match between the current utterance vector $u$ and its substructure vector $m_i$ by taking their inner product followed by a softmax.
\begin{equation}
p_i = \text{softmax}(u^T m_i),
\end{equation}
where $\text{softmax}(z_i) = e^{z_i} / \sum_j e^{z_j}$ and $p_i$ can be viewed as attention distribution for modeling important substructures from external knowledge in order to understand the current utterance.
%For example, 

\paragraph{Sentence Representation}
In order to encode the knowledge-guided structure, 
%the memory vectors are summed 
a vector $h$ is a sum over the encoded knowledge embeddings weighted by the attention distribution.
\begin{equation}
h = \sum_i p_i m_i,
\end{equation}
which indicates that the sentence pays different attention to different substructures guided from external knowledge.
Because the function from input to output is smooth, we can easily compute gradients and back propagate through it.
%Other recently proposed forms of memory or attention take this approach, Bahdanau et al. [2] and Graves et al. [8], see also [9].
Then the sum of the substructure vector $h$ and the current input embedding $u$ are then passed through a neural network model $M_\text{out}$ to generate an output knowledge-guided representation $o$.
\begin{equation}
\label{eq:encoding}
o = M_\text{out}(h + u),
\end{equation}
where we employ a fully-connected dense network for $M_\text{out}$.

\paragraph{Sequence Tagging}
To estimate the tag sequence $\vec{y}$ corresponding to an input word sequence $\vec{s}$, we use an RNN module for training a slot tagger, where the knowledge-guided representation $o$ is fed into the input of the model in order to incorporate the structure information.
%The detail is presented in the following section.
\begin{equation}
\vec{y} = \text{RNN}(o, \vec{s})
\end{equation}

\section{Recurrent Neural Network Tagger}
\label{sec:tagger}
\subsection{Chain-Based RNN Tagger}

%The goal of the NLU model is to assign a semantic tag for each word in the current utterance.
Given $\vec{s} = w_1, ..., w_T$, the model is to predict $\vec{y} = y_1, ..., y_T$ where the tag $y_i$ is aligned with the word $w_i$.
% DILEK: The paragraph above could be redundant.
We use the Elman RNN architecture, consisting of an input layer, a hidden layer, and an output layer~\cite{elman1990finding}.
The input, hidden and output layers consist of a set of neurons representing the input, hidden, and output at each time step $t$, $w_t$, $h_t$, and $y_t$, respectively.
%Figure~\ref{fig:rnn-tagging} illustrates the architecture of the plain RNN model.
\begin{eqnarray}
\label{eq:hidden}
h_t &=& \phi(W w_t + U h_{t-1}),\\
\label{eq:y}
\hat{y_t} &=& \text{softmax}(V h_t),
\end{eqnarray}
where $\phi$ is a smooth bounded function such as tanh, and $\hat{y_t}$ is the probability distribution over of semantic tags given the current hidden state $h_t$.
The sequence probability can be formulated as
\begin{equation}
\label{eq:obj}
p(\vec{y}\mid \vec{s}) = p(\vec{y}\mid w_1, ..., w_T) = \prod_i p(y_i\mid w_1, ..., w_i).
\end{equation}
The model can be trained using backpropagation to maximize the conditional likelihood of the training set labels.

To overcome the frequent vanishing gradients issue when modeling long-term dependencies, gated RNN was designed to use a more sophisticated activation function than a usual activation function, consisting of affine transformation followed by a simple element-wise nonlinearity by using gating units~\cite{chung2014empirical}, such as long short-term memory (LSTM) and gated recurrent unit (GRU)~\cite{hochreiter1997long,cho2014properties}.
RNNs employing either of these recurrent units have been shown to perform well in tasks that require capturing long-term dependencies~\cite{mesnil2015using,yao2014spoken,graves2013speech,sutskever2014sequence}.
In this paper, we use RNN with GRU cells to allow each recurrent unit to adaptively capture dependencies of different time scales~\cite{cho2014properties,chung2014empirical},
because RNN-GRU can yield comparable performance as RNN-LSTM with need of fewer parameters and less data for generalization~\cite{chung2014empirical} %,jozefowicz2015empirical}.

A GRU has two gates, a \emph{reset gate} $r$, and an \emph{update gate} $z$~\cite{cho2014properties,chung2014empirical}.
The reset gate determines the combination between the new input and the previous memory, and the update gate decides how much the unit updates its activation, or content.
\begin{eqnarray}
\label{eq:reset}
r &=& \sigma( W^r w_t + U^r h_{t-1}),\\
\label{eq:update}
z &=& \sigma( W^z w_t + U^z h_{t-1}),
\end{eqnarray}
where $\sigma$ is a logistic sigmoid function.

Then the final activation of the GRU at time $t$, $h_t$, is a linear interpolation between the previous activation $h_{t-1}$ and the candidate activation $\tilde{h_t}$:
\begin{eqnarray}
\label{eq:candidate_h}
h_t &=& (1 - z) \odot \tilde{h_t} + z \odot h_{t-1}, \\
\tilde{h_t} &=& \phi( W^h w_t + U^h (h_{t-1}\odot r) )),
\end{eqnarray}
where $\odot$ is an element-wise multiplication.
When the reset gate is off,
it effectively makes the unit act as if it is reading the first symbol of an input sequence,
allowing it to forget the previously computed state.
%Figure~\ref{fig:gru} shows the gating mechanism of a GRU cell.
Then $\hat{y_t}$ can be computed by (\ref{eq:y}).

\subsection{Knowledge-Guided RNN Tagger}
In order to model the encoded knowledge from previous turns, for each time step $t$, the knowledge-guided sentence representation $o$ in (\ref{eq:encoding}) is fed into the RNN model together with the word $w_t$.
For the plain RNN, the hidden layer can be formulated as
\begin{equation}
h_t = \phi(M o + W w_t + U h_{t-1})
\end{equation}
to replace (\ref{eq:hidden}) as illustrated in the right block of Figure~\ref{fig:model}.
RNN-GRU can incorporate the encoded knowledge in the similar way, where $M o$ can be added into gating mechanisms for modeling contextual knowledge similarly.

\begin{figure}[t]
\vspace{-2mm}
\centering
\includegraphics[width=\linewidth]{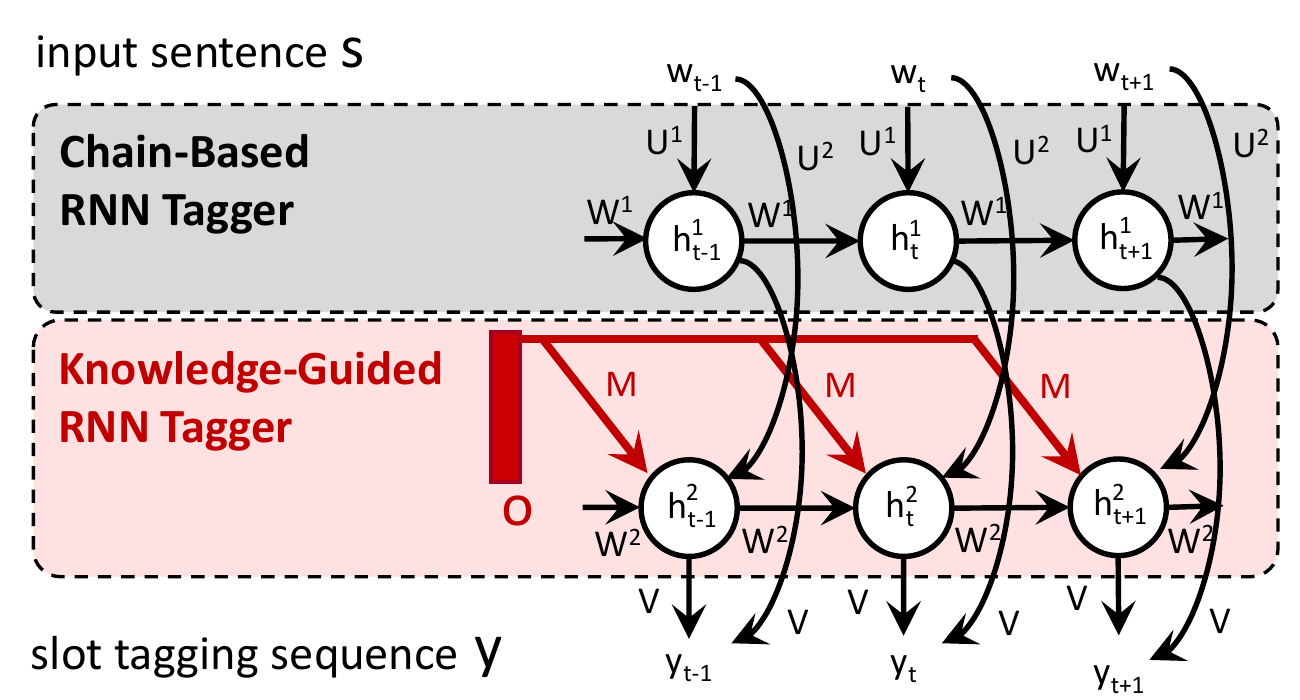}
\vspace{-5mm}
\caption{The joint tagging model that incorporates a chain-based RNN tagger (upper block) and a knowledge-guided RNN tagger (lower block).}
\vspace{-3mm}
\label{fig:joint_tagger}
\end{figure}

\subsection{Joint RNN Tagger}
Because the chain-based tagger and the knowledge-guided tagger carry different information, the joint RNN tagger is proposed to balance the information between two model architectures.
Figure~\ref{fig:joint_tagger} presents the architecture of the joint RNN tagger.
\begin{eqnarray}
h_t^1 &=& \phi(W^1 w_t + U^1 h_{t-1}),\\
h_t^2 &=& \phi(M o + W^2 w_t + U^2 h_{t-1}),\\
\hat{y}_t &=& \text{softmax}(V(\alpha h_t^1 + (1 - \alpha) h_t^2)),
\end{eqnarray}
where $\alpha$ is the weight for balancing chain-based and knowledge-guided information.
By jointly considering chain-based information ($h_t^1$) and knowledge-guided information ($h_t^2$), the joint RNN tagger is expected to achieve better generalization, and the performance may be less sensitive to poor structures from external knowledge. In the experiments, $\alpha$ is set to $0.5$ for balancing two sides.
The objective of the proposed model is to maximize the sequence probability $p(\vec{y}\mid \vec{s})$ in (\ref{eq:obj}), and the model can be trained in an end-to-end manner, where the error would be back-propagated through the whole architecture.

\section{Experiments}
\label{sec:exp}

\begin{table*}[t]
    \centering
  \begin{tabular}{llclccc}
\hline
\multirow{2}{*}{} & \multicolumn{3}{c}{Model} & \multicolumn{3}{c}{Dataset} \\
\cline{2-7}
& Encoder ($M_\text{kg}$/$M_\text{in}$) & Knowledge & Tagger & Small & Medium & Large\\
\hline\hline
%Baseline & - & \xmark & RNN & 72.59	& 85.06 & 92.69 \\
Baseline & - & \xmark & CRF & 58.94 & 78.74 & 89.73 \\
 & - & \xmark & RNN & 68.58 & 84.55 & 92.97 \\
%& - & \xmark & RNN-LSTM  & 70.97 & 86.47 & 93.14 \\
%& - & \xmark & RNN-BLSTM & 68.37 & 86.66 & 93.47 \\
& CNN & \xmark & RNN & 73.57 & 85.52 & 93.88\\
\hline
Structural & - & \cmark & CRF & 59.55 & 78.71 & 90.13\\
& DCNN & \cmark & RNN & 70.24 & 83.80 & 93.25\\
& Tree-RNN & \cmark & RNN & 73.50  & 83.92 & 92.28 \\
\hline
Proposed & K-SAN (NN) & \cmark & RNN & 74.11$^\dag$ & 85.97 & 93.98$^\dag$ \\
& K-SAN (RNN) & \cmark & RNN & 73.13 & 86.85$^\dag$ & \bf 94.97$^\dag$ \\
& K-SAN (CNN) & \cmark & RNN & \bf 74.60$^\dag$ & \bf 87.99$^\dag$ & 94.86$^\dag$ \\
%\midrule
%\multicolumn{4}{l}{Fine-Chosen Model Integration} & 76.01 & 87.62 & 95.17\\
\hline
  \end{tabular}
  \caption{The F1 scores of predicted slots on the different size of ATIS training examples, where K-SAN utilizes the dependency relations parsed from the Stanford parser. Small: 1/40 set; Medium: 1/10 set; Large: original set. ($\dag$ indicates that the performance is significantly better than all baseline models with $p<0.05$ in the t-test.)}
  \label{tab:result}
\vspace{-3mm}
\end{table*}

\subsection{Experimental Setup}
The dataset for experiments is the benchmark ATIS corpus, which is extensively used by the NLU community~\cite{mesnil2015using}.
There are 4978 training utterances selected from Class A (context independent) in the ATIS-2 and ATIS-3, while there are 893 utterances selected from the ATIS-3 Nov93 and Dec94.
In the experiments, we only use lexical features.
In order to show the robustness to data scarcity, we conduct the experiments with 3 different sizes of training data (Small, Medium, and Large), where Small is 1/40 of the original set, Medium is 1/10 of the original set, and Large is the full set.
The evaluation metrics for NLU is F-measure on the predicted slots\footnote{The used evaluation script is \texttt{conlleval}.}.

For experiments with K-SAN, we parse all data with the Stanford dependency parser~\cite{chen2014fast} and represent words as their embeddings trained on the in-domain data, where the parser is pre-trained on PTB.
The loss function is cross-entropy, and the optimizer we use is adam with the default setting~\cite{kingma2014adam}, where the learning rate $\lambda=0.001$, $\beta_1=0.9$, $\beta_2=0.999$, and $\epsilon=1e^{-08}$.
The maximum iteration for training our K-SAN models is set as 300.
%, and the number of dimensions is 150 for the embedding layer and 50 for all hidden layers in the model.
The dimensionality of input word embeddings is 100, and the hidden layer sizes are in $\{50, 100, 150\}$.
The dropout rates are set as $\{0.25, 0.50\}$.
All reported results are from the joint RNN tagger, and the hyperparameters are tuned in the dev set for all experiments.

\subsection{Baseline}

To validate the effectiveness of the proposed model, we compare the performance with the following baselines.
\begin{itemize}
\vspace{-1mm}
\item Baseline:
\begin{itemize}
\vspace{-1mm}
\item CRF Tagger~\cite{tur2010left}: predicts a semantic slot for each word with a context window (size = 5).
\vspace{-1mm}
\item RNN Tagger~\cite{mesnil2015using}: predicts a semantic slot for each word.
\vspace{-1mm}
\item CNN Encoder-Tagger~\cite{kim2014convolutional}: tag semantic slots with consideration of sentence embeddings learned by a convolutional model.
\vspace{-1mm}
\end{itemize}
\vspace{-1mm}
\item Structural: The NLU models utilize linguistic information when tagging slots, where DCNN and Tree-RNN are the state-of-the-art approaches for embedding sentences with linguistic structures.
\begin{itemize}
\vspace{-1mm}
\item CRF Tagger~\cite{tur2010left}: predicts slots based on the lexical (5-word window) and syntactic (dependent head in the parsing tree) features.
\vspace{-1mm}
\item DCNN~\cite{ma2015dependency}: predicts slots by incorporating sentence embeddings learned by a convolutional model with consideration of dependency tree structures.
\vspace{-1mm}
\item Tree-RNN~\cite{tai2015improved}: predicts slots with sentence embeddings learned by an RNN model based on the tree structures of sentences.
\vspace{-1mm}
\end{itemize}
\vspace{-1mm}
\end{itemize}

\begin{figure*}[t]
\vspace{-1mm}
\centering
\includegraphics[width=0.92\linewidth]{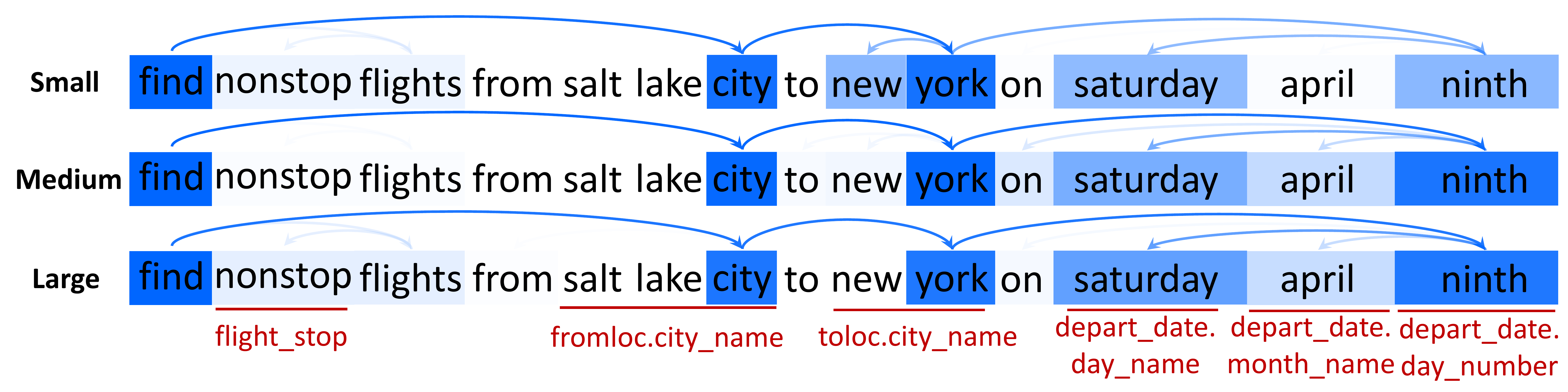}
\vspace{-3mm}
\caption{The visualization of the decoded knowledge-guided structural attention for both {\bf relations} and {\bf words} learned from different size of training data. Relations and words with darker color indicate higher attention weights generated by the proposed K-SAN with CNN. The slot tags are shown in the figure for reference.
Note that the dependency relations are incorrectly parsed by the Stanford parser in this example, but our model is still able to benefit from the structural information.}
\vspace{-2mm}
\label{fig:attention}
\end{figure*}

\subsection{Slot Filling Results}
Table~\ref{tab:result} shows the performance of slot filling on different size of training data, where there are three datasets (Small, Medium, and Large use 1/40, 1/10, and whole training data).
For baselines (models without knowledge features), CNN Encoder-Tagger achieves the best performance on all datasets.
%The probable reason is that BLSTM requires enough training data to learn more parameters for gates.

Among structural models (models with knowledge encoding), Tree-RNN Encoder-Tagger performs better for Small data but slightly worse than the DCNN Encoder-Tagger.

CNN~\cite{kim2014convolutional} performs better compared to DCNN~\cite{ma2015dependency} and Tree-RNN~\cite{tai2015improved}, even though CNN does not leverage external knowledge when encoding sentences.
When comparing the NLU performance between baselines and other state-of-the-art structural models, there is no significant difference.
This suggests that encoding sentence information without distinguishing substructure may not capture salient semantics in order to improve understanding performance.

Among the proposed K-SAN models, CNN for encoding performs best on Small (75\% on F1) and Medium (88\% on F1), and RNN for encoding performs best on the Large set (95\% on F1).
Also, most of the proposed models outperform all baselines, where the improvement for the small dataset is more significant.
This suggests that the proposed models carry better generalization and are less sensitive to unseen data.
For example, given an utterance ``{\it which flights leave on monday from montreal and arrive in chicago in the morning}'', ``{\it morning}'' can be correctly tagged with a semantic tag \textsf{B-arrive\_time.period\_of\_day} by K-SAN, but it is incorrectly tagged with \textsf{B-depart\_time.period\_of\_day} by baselines, because knowledge guides the model to pay correct attention to salient substructures.
The proposed model presents the state-of-the-art performance on the large dataset (RNN-BLSTM in baselines), showing the effectiveness of leveraging knowledge-guided structures for learning embeddings that can be used for specific tasks and the robustness to data scarcity and mismatch.

\subsection{Attention Analysis}

In order to show the effectiveness of boosting performance by learning correct attention from much smaller training data through the proposed model,
we present the visualization of the attention for both words and relations decoded by K-SAN with CNN in the Figure~\ref{fig:attention}.
The darker color of blocks and lines indicates the higher attention for words and relations respectively.
From the figure, the words and the relations with higher attention are the most crucial parts for predicting correct slots, e.g. origin, destination, and time.
Furthermore, the difference of attention distribution between three datasets is not significant; this suggests that our proposed model is able to pay correct attention to important substructures guided by the external knowledge even the training data is scarce.

\begin{figure*}[htb]
\begin{minipage}[b]{.47\linewidth}
  \centering
  \centerline{\includegraphics[width=\linewidth]{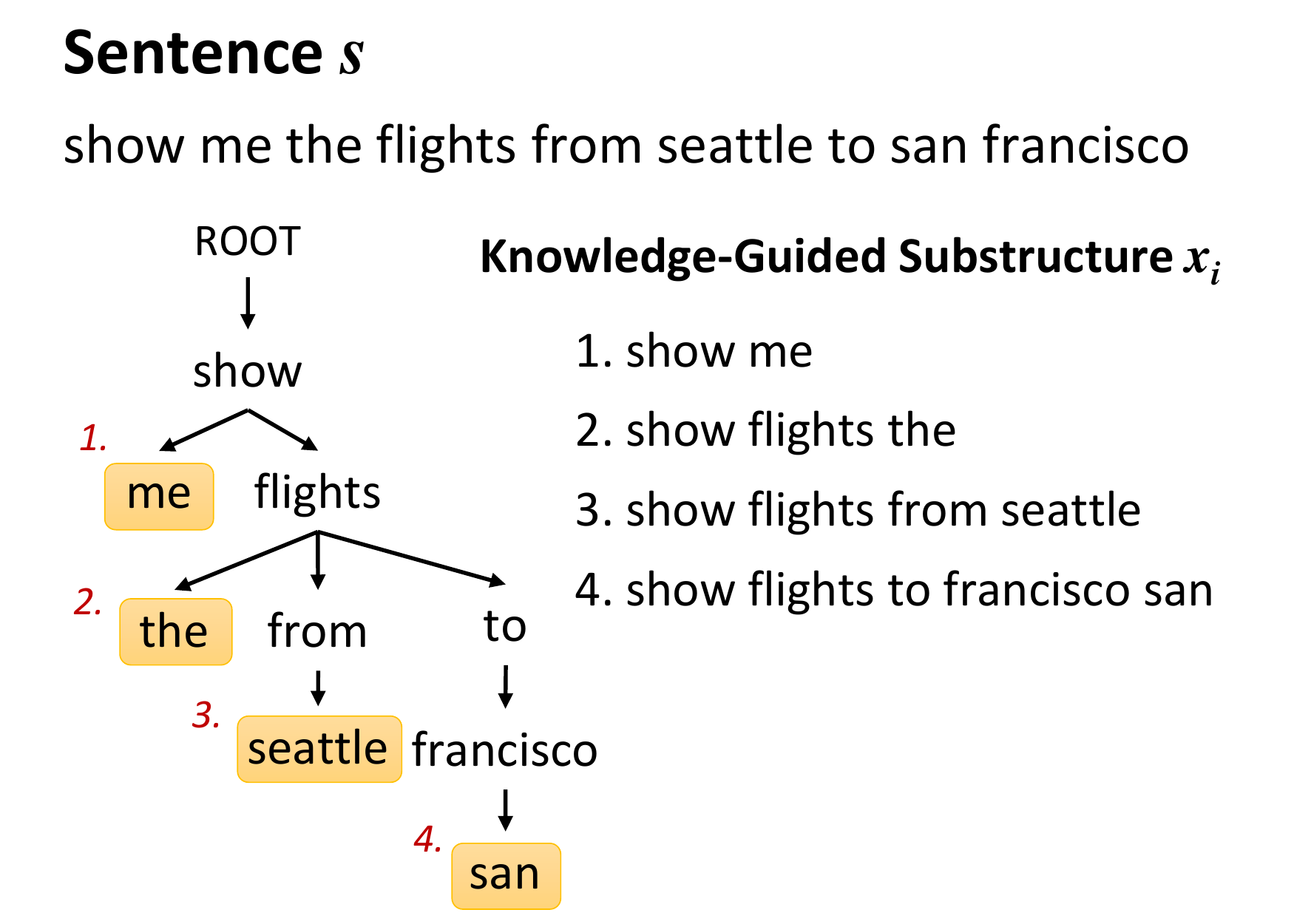}}
%  \vspace{1.5cm}
  \centerline{(a) Syntax: the dependency tree}\medskip
\end{minipage}
\hfill
\begin{minipage}[b]{0.47\linewidth}
  \centering
  \centerline{\includegraphics[width=\linewidth]{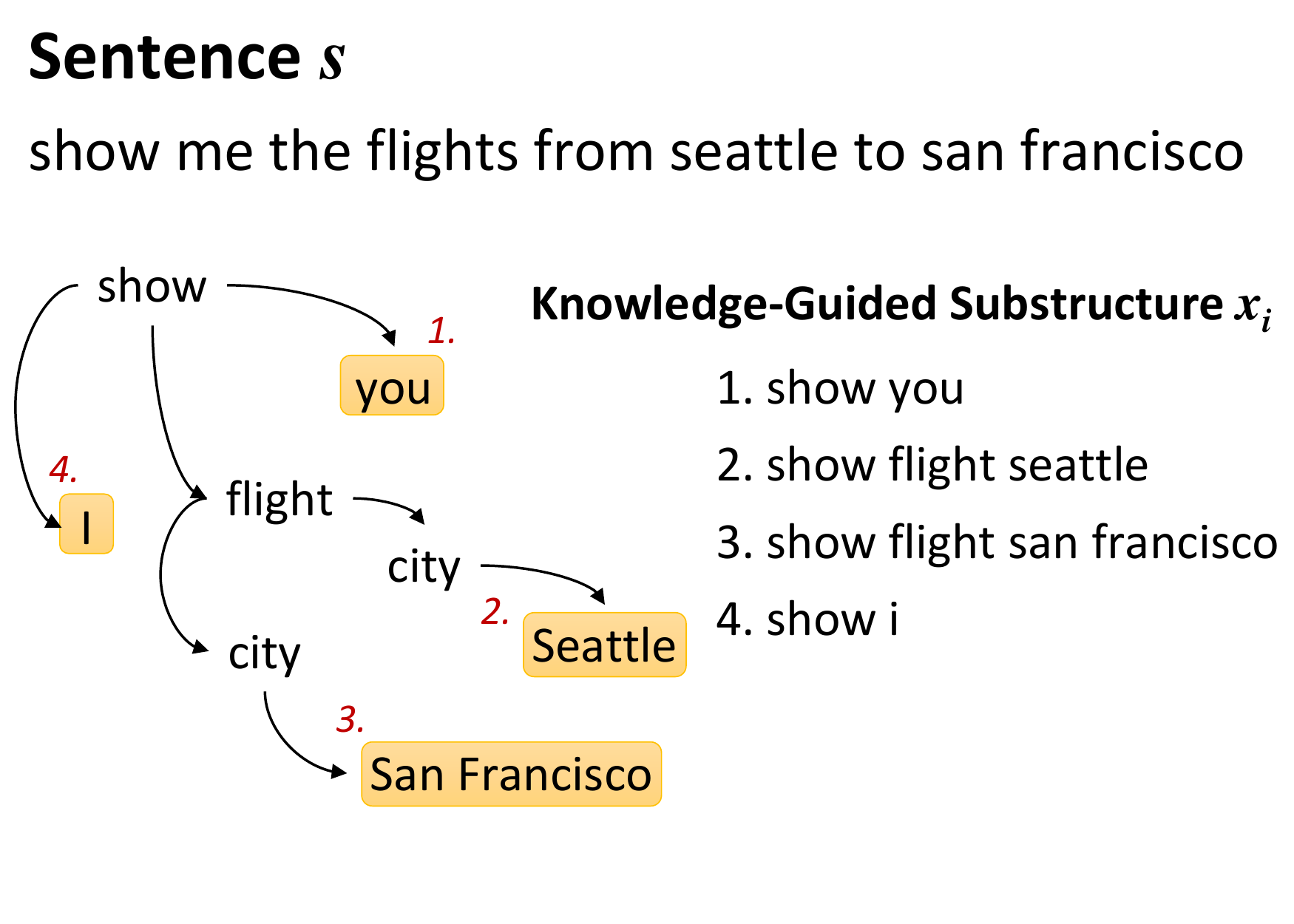}}
%  \vspace{1.5cm}
  \centerline{(b) Semantics: the AMR graph}\medskip
\end{minipage}
\vspace{-4mm}
\caption{The constructing procedure of knowledge-guided substructures, $x_i$, on an example sentence $s$.}
%\vspace{-2mm}
\label{fig:substructure}
\end{figure*}

\begin{table*}[t]
    \centering
  \begin{tabular}{cllrccc}
\hline
%\multicolumn{2}{c}{Model} & \multicolumn{3}{c}{Dataset} \\
Approach & \multicolumn{3}{c}{Knowledge (Max \#Substructure)} & Small & Medium & Large\\
\hline\hline
\multirow{4}{*}{CRF} & \multirow{2}{*}{Dependency Tree} & Stanford & - & 59.55 & 78.71 & 90.13\\
 & & SyntaxNet & - & 61.09 & 78.87 & 90.92\\
 \cline{2-7}
 & \multirow{2}{*}{AMR Graph} & Rule-Based & - & 59.55 & 79.15 & 89.97\\
 & & JAMR & - & 61.12 & 78.64 & 90.25\\
\hline
\multirow{4}{*}{K-SAN (CNN)} & \multirow{2}{*}{Dependency Tree} & Stanford & 53 & \bf 74.60 & 87.99 & 94.86 \\
& & SyntaxNet & 25 & 74.35 & \bf 88.40 & \bf 95.00\\
 \cline{2-7}
& \multirow{2}{*}{AMR Graph} & Rule-Based & 19 & 74.32 & 88.14 & 94.85\\
& & JAMR & 8 & 74.27 & 88.27 & 94.89\\
\hline
  \end{tabular}
  \vspace{-1.5mm}
  \caption{The F1 scores of predicted slots with knowledge from different resources.}
  \label{tab:parser}
\vspace{-2mm}
\end{table*}

\subsection{Knowledge Generalization}
In order to show the capacity of generalization to different knowledge resources, we perform the K-SAN model for different knowledge bases.
Below we compare two types of knowledge formats: dependency tree and Abstract Meaning Representation (AMR).
AMR is a semantic formalism in which the meaning of a sentence is encoded as a rooted, directed, acyclic graph~\cite{banarescu2013amr}, where nodes represent concepts, and labeled directed edges represent the relations between two concepts.
The formalism is based on propositional logic and neo-Davidsonian event representations~\cite{parsons1990events,davidson1967logical}.
The semantic concepts in AMR were leveraged to benefit multiple NLP tasks~\cite{liu2015toward}.
Unlike syntactic information from dependency trees, the AMR graph contains semantic information, which may offer more specific conceptual relations.
Figure~\ref{fig:substructure} shows the comparison of a dependency tree and an AMR graph associated with the same example utterance and how the knowledge-guided substructures are constructed.

Table~\ref{tab:parser} presents the performance of CRF and K-SAN with CNN taggers that utilize dependency relations and AMR edges as knowledge guidance on the same datasets, where CRF takes the head words from either dependency trees or AMR graphs as additional features and K-SAN incorporates knowledge-guided substructures as illustrated in Figure~\ref{fig:substructure}.
The dependency trees are obtained from the Stanford dependency parser or the SyntaxNet parser\footnote{\url{https://github.com/tensorflow/models/tree/master/syntaxnet}}, and AMR graphs are generated by a rule-based AMR parser or JAMR\footnote{\url{https://github.com/jflanigan/jamr}}.

Among four knowledge resources (different types and obtained from different parsers), all results show the similar performance for three sizes of datasets.
The maximum number of substructures for the dependency tree is larger than the number in the AMR graph (53 and 25 v.s. 19 and 8), because syntax is more general and may provide richer cues for guiding more attention while semantics is more specific and may offer stronger guidance.
In sum, the models applying four different resources achieve similar performance, and all significantly outperform the state-of-the-art NLU tagger, showing the effectiveness, generalization, and robustness of the proposed K-SAN model.

\section{Conclusion}
\label{sec:con}
This paper proposes a novel model, knowledge-guided structural attention networks (K-SAN), that leverages prior knowledge as guidance to incorporate non-flat topologies and learn suitable attention for different substructures that are salient for specific tasks.
The structured information can be captured from small training data, so the model has better generalization and robustness.
The experiments show benefits and effectiveness of the proposed model on the language understanding task, where all knowledge-guided substructures captured by different resources help tagging performance, and the state-of-the-art performance is achieved on the ATIS benchmark dataset.

\newpage
\bibliography{refs}

\begin{thebibliography}{}

\bibitem[\protect\citename{Banarescu \bgroup et al.\egroup
  }2013]{banarescu2013amr}
Laura Banarescu, Claire Bonial, Shu Cai, Madalina Georgescu, Kira Griffitt, Ulf
  Hermjakob, Kevin Knight, Philipp Koehn, Martha Palmer, and Nathan Schneider.
\newblock 2013.
\newblock Abstract meaning representation for sembanking.
\newblock In {\em Proceedings of the Linguistic Annotation Workshop and
  Interoperability with Discourse}.

\bibitem[\protect\citename{Celikyilmaz and
  Hakkani-Tur}2015]{celikyilmaz2015unsupervised}
Asli Celikyilmaz and Dilek Hakkani-Tur.
\newblock 2015.
\newblock Convolutional neural network based semantic tagging with entity
  embeddings.
\newblock In {\em NIPS Workshop on Machine Learning for SLU and Interaction}.

\bibitem[\protect\citename{Chelba \bgroup et al.\egroup
  }2003]{chelba2003speech}
Ciprian Chelba, Monika Mahajan, and Alex Acero.
\newblock 2003.
\newblock Speech utterance classification.
\newblock In {\em 2003 IEEE International Conference on Acoustics, Speech, and
  Signal Processing, 2003. Proceedings.(ICASSP)}, volume~1, pages I--280. IEEE.

\bibitem[\protect\citename{Chen and Manning}2014]{chen2014fast}
Danqi Chen and Christopher~D Manning.
\newblock 2014.
\newblock A fast and accurate dependency parser using neural networks.
\newblock In {\em EMNLP}, pages 740--750.

\bibitem[\protect\citename{Chen \bgroup et al.\egroup }2014]{chen2014deriving}
Yun-Nung Chen, Dilek Hakkani-Tur, and Gokan Tur.
\newblock 2014.
\newblock Deriving local relational surface forms from dependency-based entity
  embeddings for unsupervised spoken language understanding.
\newblock In {\em 2014 IEEE Spoken Language Technology Workshop (SLT)}, pages
  242--247. IEEE.

\bibitem[\protect\citename{Chen \bgroup et al.\egroup }2015]{chen2015matrix}
Yun-Nung Chen, William~Yang Wang, Anatole Gershman, and Alexander~I Rudnicky.
\newblock 2015.
\newblock Matrix factorization with knowledge graph propagation for
  unsupervised spoken language understanding.
\newblock {\em Proceedings of ACL-IJCNLP}.

\bibitem[\protect\citename{Cho \bgroup et al.\egroup }2014]{cho2014properties}
Kyunghyun Cho, Bart van Merri{\"e}nboer, Dzmitry Bahdanau, and Yoshua Bengio.
\newblock 2014.
\newblock On the properties of neural machine translation: Encoder-decoder
  approaches.
\newblock {\em arXiv preprint arXiv:1409.1259}.

\bibitem[\protect\citename{Chung \bgroup et al.\egroup
  }2014]{chung2014empirical}
Junyoung Chung, Caglar Gulcehre, KyungHyun Cho, and Yoshua Bengio.
\newblock 2014.
\newblock Empirical evaluation of gated recurrent neural networks on sequence
  modeling.
\newblock {\em arXiv preprint arXiv:1412.3555}.

\bibitem[\protect\citename{Davidson}1967]{davidson1967logical}
Donald Davidson.
\newblock 1967.
\newblock The logical form of action sentences.

\bibitem[\protect\citename{Deoras and Sarikaya}2013]{deoras2013deep}
Anoop Deoras and Ruhi Sarikaya.
\newblock 2013.
\newblock Deep belief network based semantic taggers for spoken language
  understanding.
\newblock In {\em INTERSPEECH}, pages 2713--2717.

\bibitem[\protect\citename{Elman}1990]{elman1990finding}
Jeffrey~L Elman.
\newblock 1990.
\newblock Finding structure in time.
\newblock {\em Cognitive science}, 14(2):179--211.

\bibitem[\protect\citename{Graves \bgroup et al.\egroup
  }2013]{graves2013speech}
Alan Graves, Abdel-rahman Mohamed, and Geoffrey Hinton.
\newblock 2013.
\newblock Speech recognition with deep recurrent neural networks.
\newblock In {\em 2013 IEEE International Conference on Acoustics, Speech and
  Signal Processing (ICASSP)}, pages 6645--6649. IEEE.

\bibitem[\protect\citename{Haffner \bgroup et al.\egroup
  }2003]{haffner2003optimizing}
Patrick Haffner, Gokhan Tur, and Jerry~H Wright.
\newblock 2003.
\newblock Optimizing svms for complex call classification.
\newblock In {\em 2003 IEEE International Conference on Acoustics, Speech, and
  Signal Processing, 2003. Proceedings.(ICASSP)}, volume~1, pages I--632. IEEE.

\bibitem[\protect\citename{Heck \bgroup et al.\egroup
  }2013]{heck2013leveraging}
Larry~P Heck, Dilek Hakkani-T{\"u}r, and Gokhan Tur.
\newblock 2013.
\newblock Leveraging knowledge graphs for web-scale unsupervised semantic
  parsing.
\newblock In {\em INTERSPEECH}, pages 1594--1598.

\bibitem[\protect\citename{Hochreiter and Schmidhuber}1997]{hochreiter1997long}
Sepp Hochreiter and J{\"u}rgen Schmidhuber.
\newblock 1997.
\newblock Long short-term memory.
\newblock {\em Neural computation}, 9(8):1735--1780.

\bibitem[\protect\citename{Huang \bgroup et al.\egroup
  }2013]{huang2013learning}
Po-Sen Huang, Xiaodong He, Jianfeng Gao, Li~Deng, Alex Acero, and Larry Heck.
\newblock 2013.
\newblock Learning deep structured semantic models for web search using
  clickthrough data.
\newblock In {\em Proceedings of the 22nd ACM international conference on
  Conference on information \& knowledge management}, pages 2333--2338. ACM.

\bibitem[\protect\citename{Kim}2014]{kim2014convolutional}
Yoon Kim.
\newblock 2014.
\newblock Convolutional neural networks for sentence classification.
\newblock {\em arXiv preprint arXiv:1408.5882}.

\bibitem[\protect\citename{Kingma and Ba}2014]{kingma2014adam}
Diederik Kingma and Jimmy Ba.
\newblock 2014.
\newblock Adam: A method for stochastic optimization.
\newblock {\em arXiv preprint arXiv:1412.6980}.

\bibitem[\protect\citename{Le and Mikolov}2014]{le2014distributed}
Quoc~V Le and Tomas Mikolov.
\newblock 2014.
\newblock Distributed representations of sentences and documents.
\newblock {\em arXiv preprint arXiv:1405.4053}.

\bibitem[\protect\citename{Ling \bgroup et al.\egroup }2015]{ling2015finding}
Wang Ling, Tiago Lu{\'\i}s, Lu{\'\i}s Marujo, Ram{\'o}n~Fernandez Astudillo,
  Silvio Amir, Chris Dyer, Alan~W Black, and Isabel Trancoso.
\newblock 2015.
\newblock Finding function in form: Compositional character models for open
  vocabulary word representation.
\newblock {\em arXiv preprint arXiv:1508.02096}.

\bibitem[\protect\citename{Liu \bgroup et al.\egroup }2013]{liu2013query}
Jingjing Liu, Panupong Pasupat, Yining Wang, Scott Cyphers, and James Glass.
\newblock 2013.
\newblock Query understanding enhanced by hierarchical parsing structures.
\newblock In {\em Automatic Speech Recognition and Understanding (ASRU), 2013
  IEEE Workshop on}, pages 72--77. IEEE.

\bibitem[\protect\citename{Liu \bgroup et al.\egroup }2015]{liu2015toward}
Fei Liu, Jeffrey Flanigan, Sam Thomson, Norman Sadeh, and Noah~A Smith.
\newblock 2015.
\newblock Toward abstractive summarization using semantic representations.
\newblock In {\em In Proceedings of the Conference of the North American
  Chapter of the Association for Computational Linguistics: Human Language
  Technologies}, pages 1077--1086.

\bibitem[\protect\citename{Ma \bgroup et al.\egroup }2015a]{ma2015dependency}
Mingbo Ma, Liang Huang, Bing Xiang, and Bowen Zhou.
\newblock 2015a.
\newblock Dependency-based convolutional neural networks for sentence
  embedding.
\newblock In {\em Proceedings of the 53rd Annual Meeting of the Association for
  Computational Linguistics and the 7th International Joint Conference on
  Natural Language Processing}, pages 174--179.

\bibitem[\protect\citename{Ma \bgroup et al.\egroup }2015b]{ma2015knowledge}
Yi~Ma, Paul~A Crook, Ruhi Sarikaya, and Eric Fosler-Lussier.
\newblock 2015b.
\newblock Knowledge graph inference for spoken dialog systems.
\newblock In {\em 2015 IEEE International Conference on Acoustics, Speech and
  Signal Processing (ICASSP)}, pages 5346--5350. IEEE.

\bibitem[\protect\citename{McTear}2004]{mctear2004spoken}
Michael~F McTear.
\newblock 2004.
\newblock {\em Spoken dialogue technology: toward the conversational user
  interface}.
\newblock Springer Science \& Business Media.

\bibitem[\protect\citename{Mesnil \bgroup et al.\egroup }2015]{mesnil2015using}
Gr{\'e}goire Mesnil, Yann Dauphin, Kaisheng Yao, Yoshua Bengio, Li~Deng, Dilek
  Hakkani-Tur, Xiaodong He, Larry Heck, Gokhan Tur, Dong Yu, et~al.
\newblock 2015.
\newblock Using recurrent neural networks for slot filling in spoken language
  understanding.
\newblock {\em IEEE/ACM Transactions on Audio, Speech, and Language
  Processing}, 23(3):530--539.

\bibitem[\protect\citename{Mikolov \bgroup et al.\egroup
  }2013]{mikolov2013distributed}
Tomas Mikolov, Ilya Sutskever, Kai Chen, Greg~S Corrado, and Jeff Dean.
\newblock 2013.
\newblock Distributed representations of words and phrases and their
  compositionality.
\newblock In {\em Advances in neural information processing systems}, pages
  3111--3119.

\bibitem[\protect\citename{Parsons}1990]{parsons1990events}
Terence Parsons.
\newblock 1990.
\newblock Events in the semantics of english: A study in subatomic semantics.

\bibitem[\protect\citename{Pieraccini \bgroup et al.\egroup
  }1992]{pieraccini1992speech}
Roberto Pieraccini, Evelyne Tzoukermann, Zakhar Gorelov, Jean-Luc Gauvain,
  Esther Levin, Chin-Hui Lee, and Jay~G Wilpon.
\newblock 1992.
\newblock A speech understanding system based on statistical representation of
  semantics.
\newblock In {\em 1992 IEEE International Conference on Acoustics, Speech, and
  Signal Processing (ICASSP)}, volume~1, pages 193--196. IEEE.

\bibitem[\protect\citename{Ravuri and Stolcke}2015]{ravuri2015recurrent}
Suman Ravuri and Andreas Stolcke.
\newblock 2015.
\newblock Recurrent neural network and lstm models for lexical utterance
  classification.
\newblock In {\em Sixteenth Annual Conference of the International Speech
  Communication Association}.

\bibitem[\protect\citename{Roth and Lapata}2016]{roth2016neural}
Michael Roth and Mirella Lapata.
\newblock 2016.
\newblock Neural semantic role labeling with dependency path embeddings.
\newblock {\em arXiv preprint arXiv:1605.07515}.

\bibitem[\protect\citename{Rudnicky and Xu}1999]{rudnicky1999agenda}
Alexander Rudnicky and Wei Xu.
\newblock 1999.
\newblock An agenda-based dialog management architecture for spoken language
  systems.
\newblock In {\em IEEE Automatic Speech Recognition and Understanding
  Workshop}, volume~13, page~17.

\bibitem[\protect\citename{Sarikaya \bgroup et al.\egroup
  }2011]{sarikaya2011deep}
Ruhi Sarikaya, Geoffrey~E Hinton, and Bhuvana Ramabhadran.
\newblock 2011.
\newblock Deep belief nets for natural language call-routing.
\newblock In {\em 2011 IEEE International Conference on Acoustics, Speech and
  Signal Processing (ICASSP)}, pages 5680--5683. IEEE.

\bibitem[\protect\citename{Sarikaya \bgroup et al.\egroup
  }2014]{sarikaya2014application}
Ruhi Sarikaya, Geoffrey~E Hinton, and Anoop Deoras.
\newblock 2014.
\newblock Application of deep belief networks for natural language
  understanding.
\newblock {\em IEEE/ACM Transactions on Audio, Speech, and Language
  Processing}, 22(4):778--784.

\bibitem[\protect\citename{Socher \bgroup et al.\egroup
  }2014]{socher2014grounded}
Richard Socher, Andrej Karpathy, Quoc~V Le, Christopher~D Manning, and Andrew~Y
  Ng.
\newblock 2014.
\newblock Grounded compositional semantics for finding and describing images
  with sentences.
\newblock {\em Transactions of the Association for Computational Linguistics},
  2:207--218.

\bibitem[\protect\citename{Sukhbaatar \bgroup et al.\egroup
  }2015]{sukhbaatar2015end}
Sainbayar Sukhbaatar, Jason Weston, Rob Fergus, et~al.
\newblock 2015.
\newblock End-to-end memory networks.
\newblock In {\em Advances in Neural Information Processing Systems}, pages
  2431--2439.

\bibitem[\protect\citename{Sutskever \bgroup et al.\egroup
  }2014]{sutskever2014sequence}
Ilya Sutskever, Oriol Vinyals, and Quoc~V Le.
\newblock 2014.
\newblock Sequence to sequence learning with neural networks.
\newblock In {\em Advances in neural information processing systems}, pages
  3104--3112.

\bibitem[\protect\citename{Tai \bgroup et al.\egroup }2015]{tai2015improved}
Kai~Sheng Tai, Richard Socher, and Christopher~D Manning.
\newblock 2015.
\newblock Improved semantic representations from tree-structured long
  short-term memory networks.
\newblock {\em arXiv preprint arXiv:1503.00075}.

\bibitem[\protect\citename{Tur and De~Mori}2011]{tur2011spoken}
Gokhan Tur and Renato De~Mori.
\newblock 2011.
\newblock {\em Spoken language understanding: Systems for extracting semantic
  information from speech}.
\newblock John Wiley \& Sons.

\bibitem[\protect\citename{Tur \bgroup et al.\egroup }2010]{tur2010left}
Gokhan Tur, Dilek Hakkani-T{\"u}r, and Larry Heck.
\newblock 2010.
\newblock What is left to be understood in atis?
\newblock In {\em Spoken Language Technology Workshop (SLT), 2010 IEEE}, pages
  19--24. IEEE.

\bibitem[\protect\citename{Tur \bgroup et al.\egroup }2012]{tur2012towards}
Gokhan Tur, Li~Deng, Dilek Hakkani-T{\"u}r, and Xiaodong He.
\newblock 2012.
\newblock Towards deeper understanding: Deep convex networks for semantic
  utterance classification.
\newblock In {\em 2012 IEEE International Conference on Acoustics, Speech and
  Signal Processing (ICASSP)}, pages 5045--5048. IEEE.

\bibitem[\protect\citename{Wang \bgroup et al.\egroup }2005]{wang2005spoken}
Ye-Yi Wang, Li~Deng, and Alex Acero.
\newblock 2005.
\newblock Spoken language understanding.
\newblock {\em IEEE Signal Processing Magazine}, 22(5):16--31.

\bibitem[\protect\citename{Weston \bgroup et al.\egroup
  }2015]{weston2015memory}
Jason Weston, Sumit Chopra, and Antoine Bordesa.
\newblock 2015.
\newblock Memory networks.
\newblock In {\em International Conference on Learning Representations (ICLR)}.

\bibitem[\protect\citename{Xiong \bgroup et al.\egroup }2016]{xiong2016dynamic}
Caiming Xiong, Stephen Merity, and Richard Socher.
\newblock 2016.
\newblock Dynamic memory networks for visual and textual question answering.
\newblock {\em arXiv preprint arXiv:1603.01417}.

\bibitem[\protect\citename{Xu and Sarikaya}2013]{xu2013convolutional}
Puyang Xu and Ruhi Sarikaya.
\newblock 2013.
\newblock Convolutional neural network based triangular {CRF} for joint intent
  detection and slot filling.
\newblock In {\em 2013 IEEE Workshop on Automatic Speech Recognition and
  Understanding (ASRU)}, pages 78--83. IEEE.

\bibitem[\protect\citename{Yang \bgroup et al.\egroup }2014]{yang2014embedding}
Bishan Yang, Wen-tau Yih, Xiaodong He, Jianfeng Gao, and Li~Deng.
\newblock 2014.
\newblock Embedding entities and relations for learning and inference in
  knowledge bases.
\newblock {\em arXiv preprint arXiv:1412.6575}.

\bibitem[\protect\citename{Yao \bgroup et al.\egroup }2013]{yao2013recurrent}
Kaisheng Yao, Geoffrey Zweig, Mei-Yuh Hwang, Yangyang Shi, and Dong Yu.
\newblock 2013.
\newblock Recurrent neural networks for language understanding.
\newblock In {\em INTERSPEECH}, pages 2524--2528.

\bibitem[\protect\citename{Yao \bgroup et al.\egroup }2014]{yao2014spoken}
Kaisheng Yao, Baolin Peng, Yu~Zhang, Dong Yu, Geoffrey Zweig, and Yangyang Shi.
\newblock 2014.
\newblock Spoken language understanding using long short-term memory neural
  networks.
\newblock In {\em 2014 IEEE Spoken Language Technology Workshop (SLT)}, pages
  189--194. IEEE.

\end{thebibliography}
\bibliographystyle{acl2012}

\end{document}